# Hierarchical-level rain image generative model based on GAN

**Zhenyuan Liu[1]·Tong Jia[1]·Xingyu Xing[1]·Jianfeng Wu[1]·Junyi Chen[1]·**

(1. Tongji University, Shanghai 201804)

**Abstract**

Autonomous vehicles are exposed to various weather during operation, which is likely to trigger the performance limitations of the perception system, leading to the safety of the intended functionality (SOTIF) problems. To efficiently generate data for testing the performance of visual perception algorithms under various weather conditions, a hierarchical-level rain image generative model, rain conditional CycleGAN (RCCycleGAN), is constructed. RCCycleGAN is based on the generative adversarial network (GAN) and can generate images of light, medium, and heavy rain. Different rain intensities are introduced as labels in conditional GAN (CGAN). Meanwhile, the model structure is optimized and the training strategy is adjusted to alleviate the problem of mode collapse. In addition, natural rain images of different intensities are collected and processed for model training and validation. Compared with the two baseline models, CycleGAN and DerainCycleGAN, the peak signal-to-noise ratio (PSNR) of RCCycleGAN on the test dataset is improved by 2.58 dB and 0.74 dB, and the structural similarity (SSIM) is improved by 18% and 8%, respectively. The ablation experiments are also carried out to validate the effectiveness of the model tuning.

## 1 Introduction

With the prosperous development of chips, artificial intelligence, communication, and sensor technologies, autonomous vehicles have gradually become one of the developing directions in the automobile industry[1]. Autonomous vehicles have great advantages in improving access efficiency and reducing travel costs, but safety problems are also a social concern. Autonomous vehicles face a variety of safety problems, in addition to the functional safety problems caused by the failure of electrical and electronic systems[2] and the information security problems caused by accidental or malicious damage to the information system[3]. They also confront the safety of the intended functionality (SOTIF)[4] problems caused by insufficient system functions, performance limitations, and reasonably foreseeable misuse. Autonomous vehicles usually consist of three components: perception system, decision system, and control system, among which the perception system is the essential component to acquire external information accurately and timely. Therefore, the SOTIF problems of the perception system are one of the important issues that restrict autonomous vehicles coming to fruition.

The perception system usually contains various sensors such as visual sensors, LiDAR, and millimeter-wave radar to accurately acquire information about roads, traffic participants, and traffic situations around the vehicle. Visual sensors are currently one of the most commonly used sensors for autonomous vehicles because of the abundance of information acquired, low cost, and low power consumption[5]. The working process of visual sensors relies on the refraction and reflection of external light. However, autonomous vehicles are exposed to various weather conditions during operation, such as rain, snow, and fog, which seriously impact the propagation of external light[6], making it difficult for visual sensors to achieve the intended functionality. Various weather not only degrade the quality of the raw image output from the visual sensor but also reduce the subsequent detection performance[7]. To improve the performance of visual sensors under various weather, existing studies have been conducted mainly from the perspectives of improving the capability of algorithms[8] or diversifying the data in datasets[9]. Both these methods rely heavily on natural image data under various weather. However, there are few relevant datasets and their categorization is simple, which limits the potential improvement of the performance of visual sensors under various weather. As a result, it is necessary to construct an image generative model to generate high-fidelity weather images. Next, we take rain as an example to summarize the existing image generation methods.

Rain image generation methods are mainly classified into two ways: 1) Physical model-based methods: By studying the physical model of raindrops, the effect of rain on the image is obtained. Garg et al[10][11] comprehensively summarized the methods of constructing rain images based on the physical model and constructed a rain streak database to simulate the motion of raindrops in the process of descending. Thereafter, Barnum et al[12] attempted to describe the motion model of rain and snow using a frequency model to obtain the effect of rain and snow on images. Tremblay M et al[13] summarized the existing methods based on physical models and theoretical analysis, combined with deep learning to construct a rain image generative model. 2) Deep learning-based methods: The rain image generative model is constructed based on neural networks and rain images are acquired by data-driven. Liu et al[14] constructed a severe weather conditions image generative model by the double residual network. Zhu et al[15] utilized CycleGAN to learn a mapping model from sunny to rainy images using a set of unpaired sunny and rainy images. Zhu et al[16] proposed an unsupervised rain removal network, which consists of a multiscale discriminator network, a generator network, and an unsupervised rain detection module, which accelerates model convergence. In conclusion, although physical model-based methods can derive the effects of different rain intensities on images, they are very prone to ignore other features of rain, such as ground wetness, ground reflections, and clouds. Consequently, it is difficult to derive a complete impact of rain conditions on images. For the deep learning-based rain image generation methods, the existing models have low accuracy and do not include different rain intensities.

In summary, the existing image generation methods cannot satisfy both the generation of hierarchical-level rain images and the high degree of image authenticity. In this paper, we propose a hierarchical-level rain image generative model based on GAN and collect natural rain images of different grades to test and verify the model's performance. The main contributions of this paper include two aspects:

- Based on CycleGAN, we construct a three-level rain image generation model for light, medium, and heavy rain by introducing rain intensity labels into the conditional generative adversarial network.
- We collect real images under sunny, light rain, medium rain, and heavy rain conditions and construct a hierarchical-level rain image dataset, covering various target objects such as vehicles, pedestrians, and bicycles.

The rest of this paper is organized as follows. In Sect.2, we analyze the existing image generative models. In Sect.3, we present the proposed hierarchical-level rain image generative model. In Sect.4, we conduct experiments to verify the model performance, and summarize the contents of the whole paper in Sect.5.

## 2 Related works

GAN was proposed in 2014[17] and is used widely in the field of image generation. It consists of two sub-networks: the generator network and the discriminator network. The generator network learns the desired output distribution from the training data and generates the simulated data from the original input. The discriminator network discriminates whether the input data is simulated data or real data. In the adversarial process, the generator network aims to obtain the simulated data that the discriminator network cannot discriminate, while the discriminator network aims to discriminate the simulated data from real data as much as possible. The generator network and the discriminator network are trained according to dynamic game rules, and their game optimization goals (joint loss function) are as follows

$$\min(G)\max(D)J(G,D) = \mathrm{E}_{x\sim P_{data}(x)}[log[D(x)]] + \mathrm{E}_{z\sim P_z(z)}\left[log[1 - D(G(z))]\right] \quad (1)$$

where $z$ is the input noise, $x$ is the real data, $D(x)$ is the probability corresponding to the real data as the input to the discriminator network, $G(z)$ is the simulated data output when the input noise is used as the input to the generator network, $D(G(z))$ is the probability corresponding to the simulated data as the input to the discriminator network, $P_{data}(x)$ is the spatial distribution of the real data, and $P_z(z)$ is the spatial distribution of the input noise. During the training process, the real data and the input noise are input into the discriminator network and the generator network, respectively. The simulated data is obtained to update the parameters of the discriminator network and the parameters of the generator network remain unchanged. At this time, the discriminator network maximizes the game optimization goal. Then, with the parameters of the discriminator network unchanged, the output of the discriminator network is back-propagated to update the parameters of the generator network, so that the discriminator network cannot discriminate between the simulated data and the real data.

CGAN (Conditional Generative Adversarial Networks)[18] was proposed in 2014, which realizes a one-to-many mapping output for GAN. This model introduces the idea of supervised learning to traditional deep learning and introduces auxiliary information y (data labels) to the original generator and discriminator network so that the generator network can generate simulated data with auxiliary guidance. DCGAN (Deep Convolutional Generative Adversarial Network)[19] was proposed in 2015 and is one of the most important variants of GAN. It uses convolutional neural networks (CNN) instead of fully connected layers to construct the generator and discriminator network, effectively solving the unstable training problem of GAN. CycleGAN (Cycle-Consistent Generative Adversarial Networks)[20] was proposed in 2017, which realizes end-to-end learning between unpaired images. This model is widely used in style transfer between two datasets, and its network schematic diagram is shown in Figure 1. This GAN includes two sets of generator and discriminator networks. In addition to satisfying the original GAN game optimization objective (adversarial loss D_loss), it also introduces a cycle consistency loss function (C_loss), that is, the real data (Real_A) and the reconstructed data (Recon_A) must be consistent, which constrains the generator network A and generator network B to be an inverse process, solving the generation problem between unpaired data.

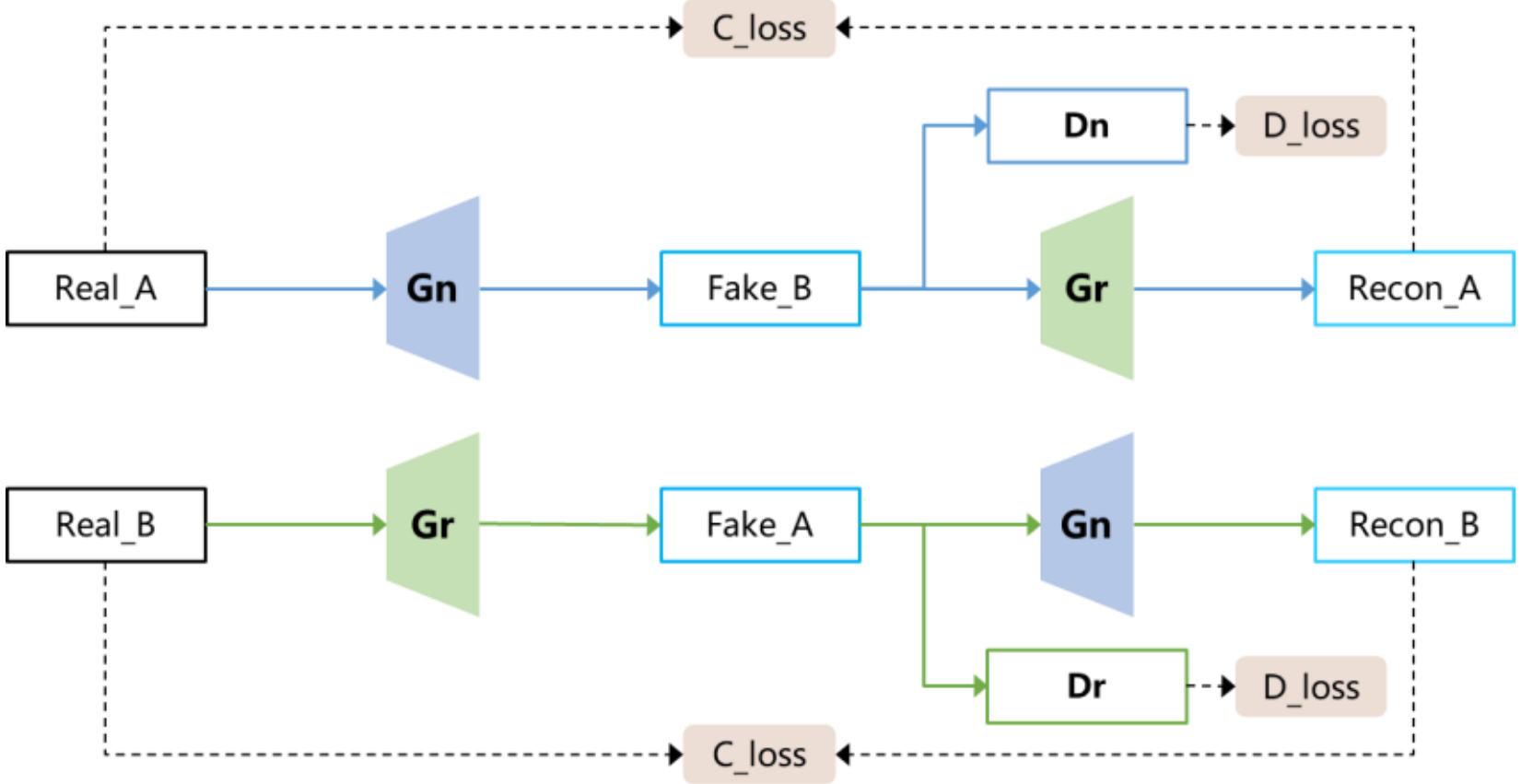

Figure 1. Schematic diagram of CycleGAN network

To summarize, GAN has been widely applied in image generation due to its simple loss function setting, but there are still problems such as unstable training, uncontrollable output, and strict training data requirements. DCGAN improves the training stability of GAN by optimizing the model structure, CGAN has the ability of one-to-many mapping output, and CycleGAN supports style transfer between unpaired images. Based on the literature[21], we construct a hierarchical-level rain image generative model with rain intensities as data labels and optimize the model details.

## 3 Hierarchical-level rain image generative model

The schematic diagram of the hierarchical-level rain image generative model (Rain Conditional CycleGAN, RCCycleGAN) is shown in Figure 2. The model comprises the following four modules:

(1) The rain mask identification network: This network is used to obtain the rain masks for both rain and simulated images, and it is an unsupervised rain mask identification module;

(2) The discriminator networks $D_n$ and $D_r$: their input is natural rain images or simulated rain images with corresponding rain intensity labels, $D_n$ is used to discriminate whether the input image belongs to a natural sunny day image or a simulated sunny day image, and $D_r$ is used to discriminate whether the input image belongs to a natural rain image or a simulated rain image;

(3) The generator networks $G_n$ and $G_r$: their input is natural rain images or simulated rain images, rain masks ,and corresponding rain intensity labels, $G_n$ is used to generate simulated sunny day images from hierarchical-level rain images (including natural and simulated rain images), and $G_r$ is used to generate simulated hierarchical-level rain images from sunny day images (including natural and simulated sunny day images);

(4) The image feature identification network: This network is used to identify the essential features of both simulated and natural images by adopting pre-trained deep convolutional networks.

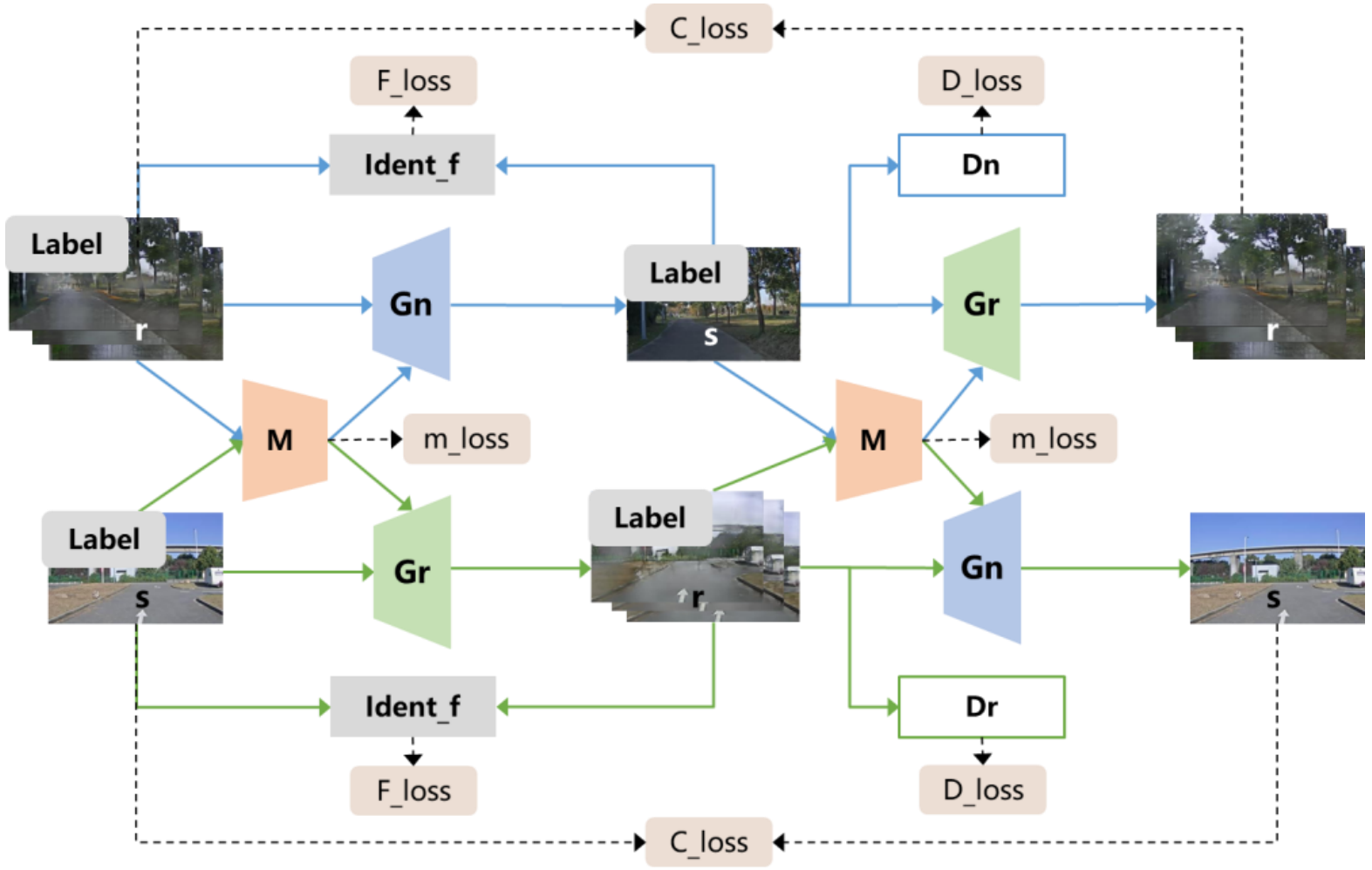

Figure 2: Schematic diagram of RCCycleGAN network

### 3.1 The rain mask identification network

In the task of converting sunny day images into rain images, the traditional CycleGAN captures the differences between the two datasets (sunny day/rain) through the discriminator network and then guides the generator network to obtain more authentic rain images. Yet, empirical evidence suggests that the information contained in sunny day and rain images is not symmetrical. Rain images are composed of background information and rain information (raindrop streaks and raindrop occlusion):

$$R = B + M \tag{2}$$

where $R$ represents the rain image; $B$ represents the background information, i.e. the sunny day image; $M$ represents the rain information, which usually consists of raindrop occlusion and raindrop streaks, i.e. the rain mask. Thus, relying solely on the CycleGAN model for this task would require an extensive amount of data and training time. Furthermore, the generator network would require considerable computational cost to complete the learning of the rain information. In the case of data scarcity and insufficient training, the CycleGAN model usually suffers from color or structure damage problems. To address these issues, some studies have proposed the design of rain information extraction modules to guide GAN to complete the learning of the rain information more accurately and quickly. Reference[21] designed a rain mask identification (RMI) network based on LSTM (Long Short Term Memory)[22], which jointly constrains the RMI process through the two branches of rain images and sunny day images. The network structure's schematic diagram is presented in Figure 3.

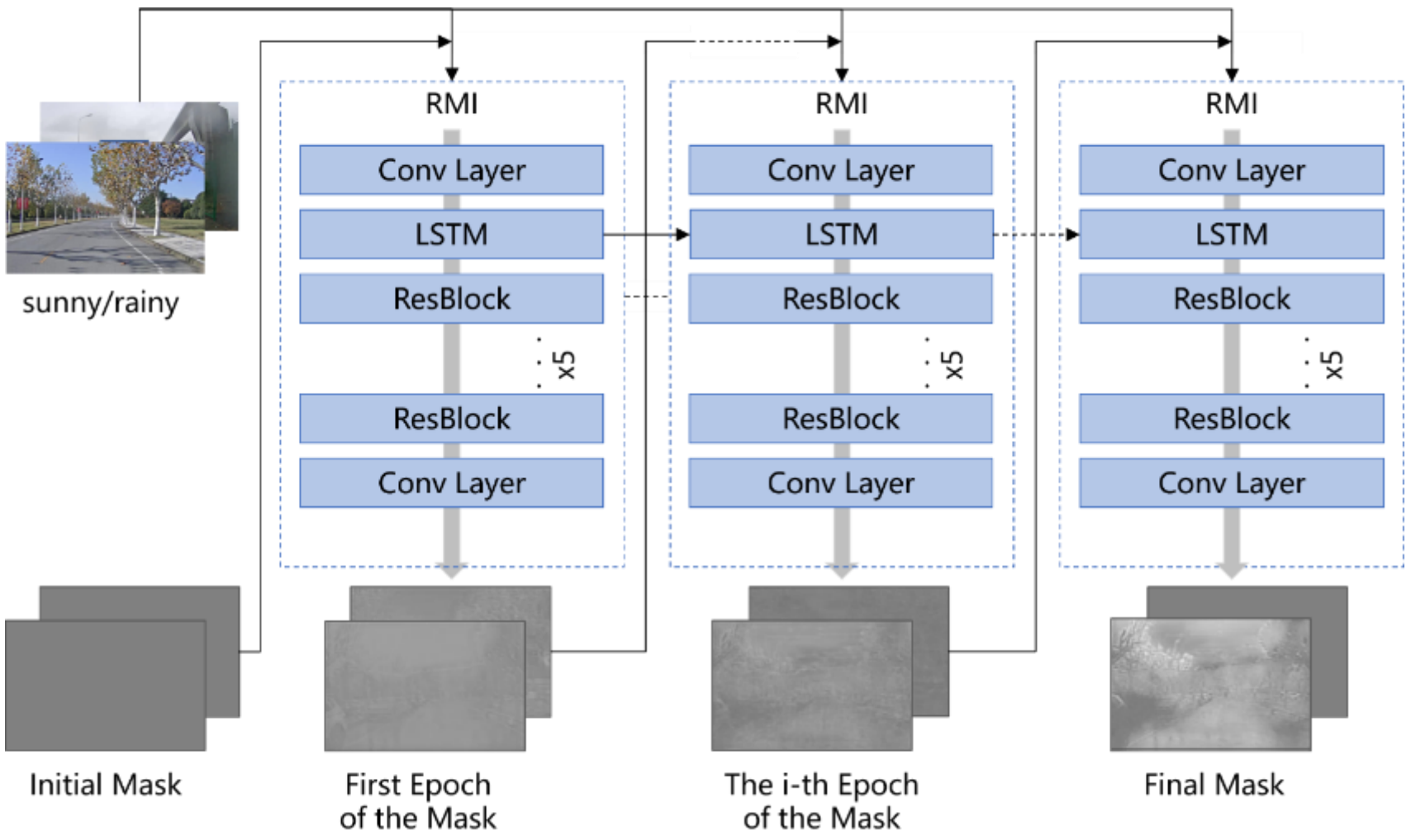

Figure 3: Schematic diagram of the rain mask identification network

The RMI network comprises several RMI modules, each of which is composed of convolutional and activation layers, LSTM, and several residual modules. First, the convolutional and activation layers extract features from the input image and the previously output rain mask. Subsequently, the LSTM module is used to update the rain mask in combination with the previous output. In addition, as the number of RMI modules increases, the number of layers in the RMI network increases dramatically, leading to the problem of gradient vanishing during network training. To address this issue, the RMI module introduces 5 residual modules, enhancing the RMI network's depth and receptive field. Finally, a single channel rain information (i.e. mask) is obtained using the convolutional layer.

To limit the information obtained by the RMI network in an unsupervised condition, we train the RMI network using prior knowledge and self-supervised learning strategies. The input for the RMI network comprises sunny day images and also rain images. For sunny day images, the mask obtained by the RMI network should be empty; for rain images, the RMI network is required to extract the all rain information, so the rain mask and the simulated rain-free image output from the generator network when superimposed need to be as close as possible to the original rain image. Specifically, its loss function is:

$$L_{Ident_m_n} = \frac{1}{N}\sum_j^N[(Ident_m(n) - Z)^2] \tag{3}$$

$$L_{Ident_m_r} = \frac{1}{N}\sum_j^N\left[\left(\left(Ident_m(r) + G_n(r)\right) - r\right)^2\right] \tag{4}$$

where, $N$ is the number of input images; $Ident_m(\cdot)$ is the RMI network; $n$ is the sunny day image; $Z$ is a distribution of the same size as the mask, but all zero; $G_n(\cdot)$ is the sunny day image generator network; $r$ is the rain image.

### 3.2 The generator network

Generator networks $G_n$ and $G_r$ adopt the same U-net structure, and their network schematic diagram is shown in Figure 4. The input of the generator network (5 channels) includes the original image (RGB 3 channels), the rain intensity label (1 channel), and the rain mask (grayscale image 1 channel), where the rain intensity label is the condition for the CGAN. The generator network is mainly composed of an encoder, a transformer, and a decoder. The encoder is composed of 4 layers of CNN, which completes the feature extraction of the input information and obtains better global information.

The transformer is composed of 10 layers of CNN, realizing rain/no-rain style transfer. The decoder is composed of 3 layers of CNN, which decodes the image features after style transfer and outputs the simulated image.

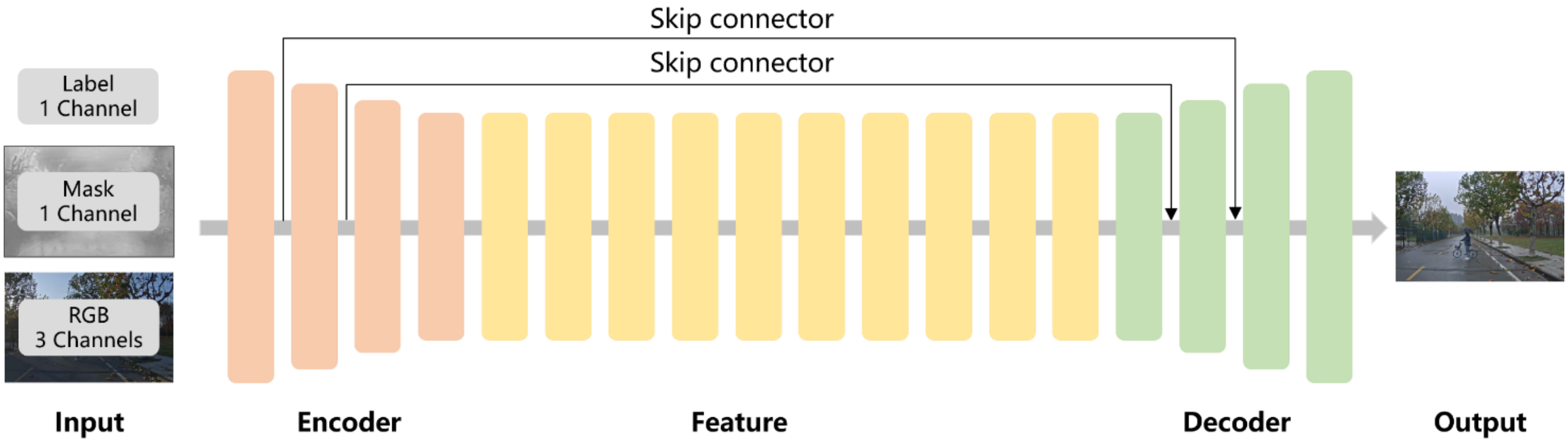

Figure 4 Schematic diagram of the generator network structure

The training goal of the generator network is to output simulated data that the discriminator cannot discriminate. This paper relies on LSGAN (Least Squares Generative Adversarial Networks) to replace the cross-entropy loss function in the traditional GAN with the least squares loss function. The generator network uses the least squares loss function to update the network model parameters, improve the stability of the GAN model, and speed up the convergence of the model. The loss function calculation method is as follows:

$$L_{Gen} = \frac{1}{3} \cdot \frac{1}{N} \sum_i^3 \sum_j^N \left[ \left( D\left( G\left(Real_j\right)\right)[i] - I_{Real} \right)^2 \right] \tag{5}$$

where $i$ is the $i$th scale output of the discriminator network; $N$ represents the number of input images; $D(\cdot)$ is the discriminator network; $G(\cdot)$ is the generator network; $Real_j$ is the jth real image in the batch; $I_{Real}$ is the ideal output of the real image input to the discriminator network, and all values in the patch are 1.

In addition, the model also retains the cycle consistency loss of CycleGAN. There are two mapping processes in this model: $G_n: R \rightarrow N$ and $G_r: N \rightarrow R$. The real data is input into the model, and after these two mapping functions, the output should be as close as possible to the original real output. Therefore, the loss function is as follows:

$$L_{Cycle} = \frac{1}{N} \sum_j^N \left[ \left\| r_{real} - G_r\left(G_n\left(r_{real}\right)\right) \right\|_1 + \left\| n_{real} - G_n\left(G_r\left(n_{real}\right)\right) \right\|_1 \right] \tag{6}$$

where $N$ denotes the number of input images; $G_n(\cdot)$ is the sunny day image generator network; $G_r(\cdot)$ is the rain image generator network; $r_{real}$ is the natural rain image; $n_{real}$ is the natural sunny day image.

### 3.3 The discriminator network

The discriminator network adopts the same multi-scale convolutional network. The input is simulated or natural images (RGB 3 channels) and the corresponding rain intensity label (1 channel), where the rain intensity label is the condition for the CGAN. Compared with the traditional approach that uses a fully connected layers as the final output of the discriminator network, the final output of the PathGAN structure is the local receptive field, which can preserve more image features and better guide the output of the generator network. The network structure is shown in Figure 5. It consists of three deep CNN with different scales obtained by averaging the pooling layers. The features in the receptive field are extracted by each convolutional layer through the convolutional kernel, and the features and structural information of the input image are extracted by the multilayer convolutional network.

Similarly, the discriminator network adopts the least squares loss function to improve the quality of the generated images and to optimize the training process of the GAN. The loss function is as follows:

$$L_{Dis} = \frac{1}{3} \cdot \frac{1}{N} \sum_i^3 \sum_j^N \left[ \left( D(Real_j)[i] - I_{Real} \right)^2 + \left( D(Fake_j)[i] - I_{Fake} \right)^2 \right] \tag{7}$$

where $i$ is the $i$-th output of the discriminator network; $N$ is the number of input images; $D(\cdot)$ is the discriminator network; $Real_j$ is the $j$-th real image in the batch; $Fake_j$ is the $j$-th simulated image in the batch; $I_{Real}$ is the ideal output when the real image is input into the discriminator network, and all values in the patch are 1; $I_{Fake}$ is the ideal output when the simulated image is input into the discriminator network, and all values in the patch are 0.

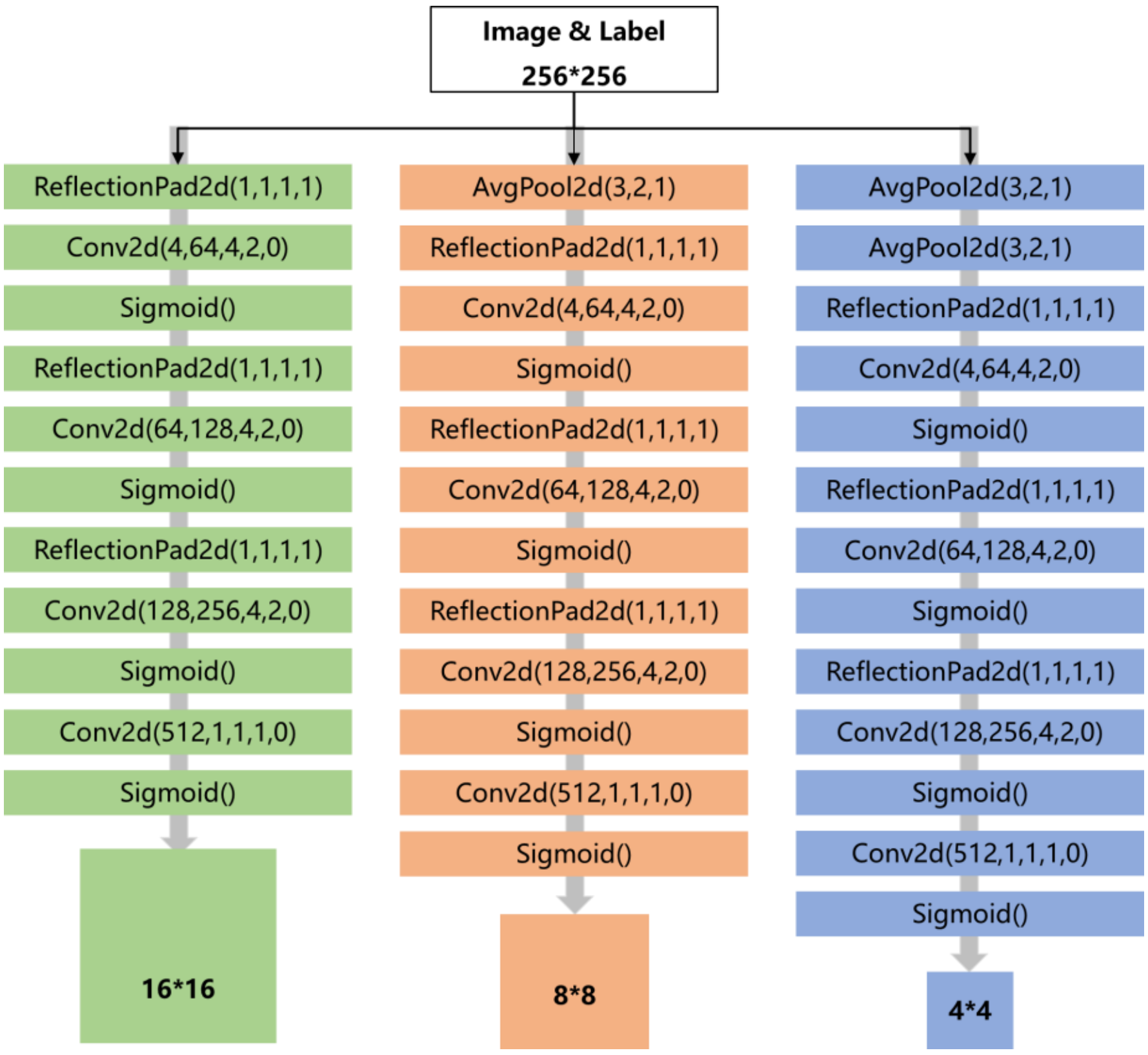

Figure 5 The structure diagram of the discriminator network

### 3.4 The image feature recognition network

The image feature identification network is a stable pre-trained VGG16 network, which extracts the features and structural information contained in real and simulated images and requires no further training. According to the prior knowledge, the simulated rain image output from the generator network is superimposed with the rain mask on the natural image. In general, the features and structural information of the image are not changed significantly. Therefore, the error between the simulated image output by the generator network and the real image after passing through the image feature identification network should be within a certain range, and its loss function is defined as follows:

$$L_{Ident_f} = \frac{1}{N} \sum_j^N \left[ \left( \left( Ident_m(r) + G_n(r) \right) - r \right)^2 \right] \tag{8}$$

### 3.5 Overall loss function

The overall loss function of the hierarchical-level rain image generative model in this paper is as follows:

$$L_{total} = \lambda_D L_{Dis} + \lambda_G L_{Gen} + \lambda_{Cycle} L_{Cycle} + \lambda_{Im} L_{Ident_m} + \lambda_{If} L_{Ident_f} \tag{9}$$

where, $L_{Ident_m} = L_{Ident_m_n} + L_{Ident_m_r}$, $\lambda_D$, $\lambda_G$, $\lambda_{Cycle}$, $\lambda_{Im}$ and $\lambda_{If}$ are the weights of each loss.

## 4 Experiment

Table 1 displays the hyperparameter setting during the training of the hierarchical-level rain image generative model. Due to the large size of the network model, the image size is set to 256*256 and the batch is set to 1. The network model uses the Adam optimizer. The generator network and the discriminator network use different learning rates to suppress the performance of the discriminator network. The generator network's initial learning rate is set at 0.0001, while the discriminator network's initial learning rate is set at 0.00008. The model undergoes 400 epochs, where its learning rate is decreased to half of the original rate at the 200th epoch and further reduced to one-fourth of the original rate at the 300th epoch. The RMI network consists of 6 RMI modules.

In Eq. (9), the loss weight coefficients are $\lambda_D$ =1、$\lambda_G$=1、$\lambda_{Cycle} = 10$、$\lambda_{Im} = 0.1$ and $\lambda_{If} = 10$. In addition, to suppress the performance of the discriminator network, we adopt a discriminator network suppression strategy, that is, the generator network parameters update every three times, while the discriminator network parameters update once.

**Table 1** Hyperparameter setting

| Hyperparameter | Value |
|---|---|
| Image_size | 256*256 |
| epoch | 400 |
| batch | 1 |
| optimizer | Adam |
| epoch_decay | 200 |
| n_RMI | 6 |
| lr_Dis | 0.0001 |
| lr_Gen | 0.00008 |

## 4.1 Dataset and evaluation metrics

The majority of current datasets are collected under good weather and lack natural rain images. A few rain datasets provide data labels with only high-level classification (with/without rain), or are artificially synthesized data with low authenticity. Additionally, some datasets are continuous scenes with little diversity. Thus, the existing datasets cannot support the training and validation of our model. The comparison of existing datasets is shown in Table 2.

**Table 2** Comparison of existing datasets.

| Dataset | Whether or not there is a rain label | Labels type | Natural or not | Paired or unpaired | Scenario diversity | Quantity (in thousands) |
|---|---|---|---|---|---|---|
| KITTI[23] | without | / | natural | unpaired | high | 15 |
| Nuscense[24] | without | / | natural | unpaired | high | 40 |
| BDD100K[25] | without | / | natural | unpaired | high | 100 |
| Cityscapes[26] | without | / | natural | unpaired | high | 25 |
| CADC[27] | without | / | natural | unpaired | high | 7.50 |
| Boreas[28] | with | with/without rain | natural | unpaired | medium | about 880 |
| SPA[29] | with | with/without rain | natural | paired | low | 0.15 |
| SIRR[30] | with | with/without rain | natural | unpaired | low | 1 |
| Rain800[31] | with | with/without rain | not(synthetic) | paired | medium | 0.80 |
| Rain12600[32] | with | 14 types of rain | not(synthetic) | paired | medium | 12.60 |
| **NMRD (ours)** | **with** | **Sunny/heavy rain/ medium rain/light rain** | **natural** | **partial paired** | **medium** | **1.28** |

To address this issue, we construct a natural multilevel rain dataset (NMRD) to support the research. Apart from generating rain images in this research, the dataset can also be utilized for natural rain removal problems. Figure 6 shows the data collection platform. It is equipped with a Delphi IFV 300 (Mobileye EyeQ3) camera to capture real road scenes. Table 3 illustrates the sensor-related parameters. Moreover, to eliminate the non-rainy condition interference, the raw images obtained from the data collection platform are pre-processed, which mainly includes the following two aspects: 1) Manually remove images where information is lost caused by raindrops completely covering the camera surface. 2) Crop to remove non-road scene information (such as the front of the vehicle) in the image, and the size of the cropped image is 1080*700.

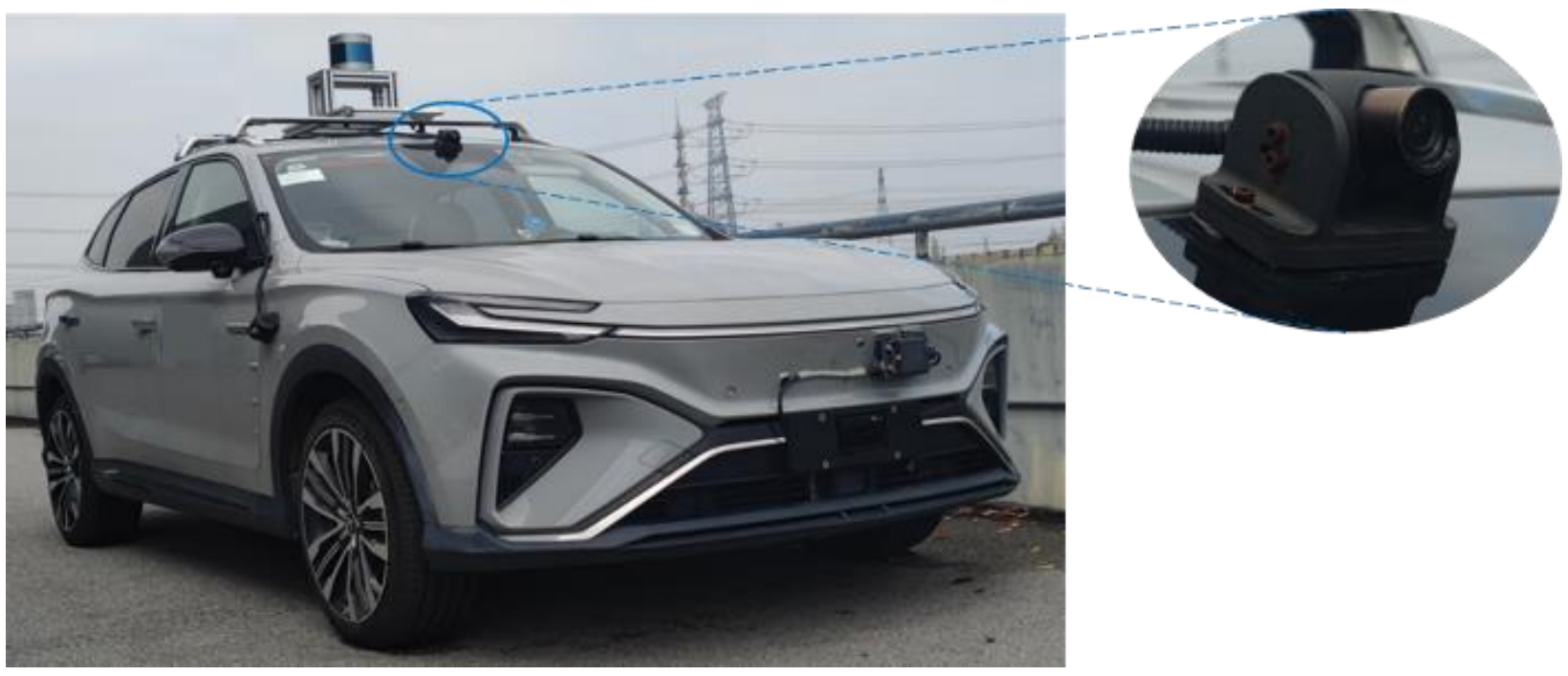

Figure 6. Data collection platform

Table 3 Parameters of the IFV 300 (Mobileye EyeQ3)

| **Camera Model** | **IFV 300（Mobileye EyeQ3）** |
|---|---|
| Resolution | 1920*1080 |
| Frame rate | 30FPS |
| Sensor type | Aptina AR0132 |
| Horizontal field of view | 52° |
| Vertical field of view | 39° |
| Dynamic range | 115dB |
| Built-in chip | Mobileye EQ3 |

The training dataset comprises four label categories – sunny day, light rain, medium rain, and heavy rain - provided by real-time weather forecasts during collection. Figure 7 shows the differences in images under different rain intensities, including pedestrians, non-motorized vehicles, and motor vehicles. Table 4 presents basic information about the dataset.

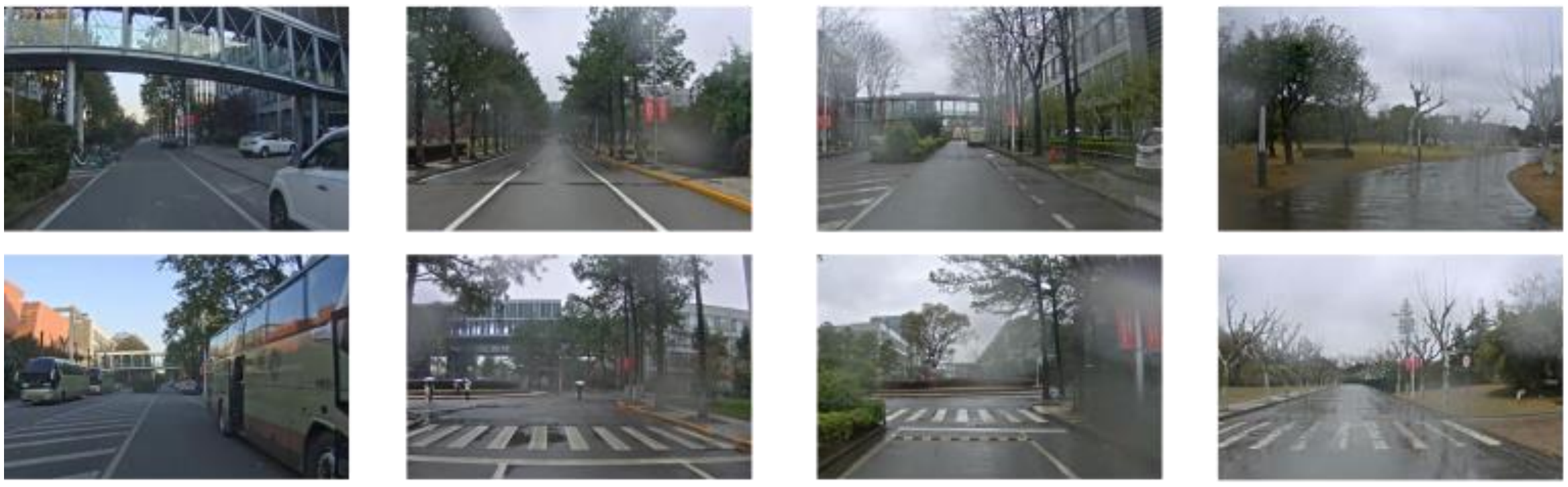

Figure 7. The NMRD dataset

Table 4 Introduction to the NMRD dataset

| | **Sunny day** | **Light rain** | **Medium rain** | **Heavy rain** | **Role** |
|---|---|---|---|---|---|
| Training data | 173 | 186 | 248 | 193 | model training |
| Test data | 24 | 24 | / | / | 24 pairs of images for quantitative evaluation of the model performance |
| Validation data | 432 | / | / | / | part of the data for qualitative validation of the model performance |
| Total | 629 | 210 | 248 | 193 | / |

Objective and unified evaluation methods for image generation effects are still absent. Most of the existing studies evaluated the image generation effects by assessing the image quality . Our test data consists of 24 paired rain images, so we use the full-reference evaluation metrics to quantify the differences between the generated rain images and the natural rain images. The existing full-reference evaluation metrics mainly assess image quality from two dimensions: pixel level and structure level. Thus, we choose peak signal-to-noise ratio (PSNR) and structural similarity (SSIM) as evaluation metrics to objectively evaluate the image generation effects.

PSNR is a commonly used full-reference evaluation index that calculates the pixel-level difference between the output image and the reference image to evaluate the degree of distortion of the generated image. The calculation method for PSNR is as follows.

$$MSE = \frac{1}{r \cdot c} \sum_{i=0}^{r-1} \sum_{j=1}^{c-1} [I_G(i,j) - I_o(i,j)]^2 \quad (10)$$

$$PSNR = 10 \cdot log_{10} \left( \frac{(2^n - 1)^2}{MSE} \right) \quad (11)$$

where $MSE$ is the Mean Square Error (MSE) of the image, $r$ is the height of the image, $c$ is the width of the image, $I_G(i,j)$ is the value of the pixel of the generated image at position $(i,j)$, $I_o(i,j)$ is the value of the pixel of the original reference image at position $(i,j)$, and $n$ is the number of bits of the image, usually 8 bits, i.e. the number of grey levels of the pixel is 256. PSNR is measured in dB, with higher values indicating less distortion in the generated image and less pixel-level error between the generated image and the reference image.

SSIM is a commonly used full-reference evaluation index for measuring the structural similarity of two images. The index evaluates the image quality from three dimensions: luminance, contrast, and structure, focusing on the correlation between neighboring pixels. The calculation method is as follows:

$$SSIM = \frac{(2\mu_x\mu_y + C_1)(2\sigma_{xy} + C_2)}{(\mu_x^2 + \mu_y^2 + C_1)(\sigma_x^2 + \sigma_y^2 + C_2)} \quad (12)$$

where $\mu_x$ is the pixel mean of the image $x$; $\mu_y$ is the pixel mean of the image $y$; $\sigma_x$ is the pixel variance of the image $x$; $\sigma_y$ is the pixel variance of the image $y$; $\sigma_{xy}$ is the covariance of the image $x$ and y;$C_1$ and $C_2$ are constants.

## 4.2 Model comparison

24 sets of paired sunny and light rain images are selected from the test data as input. Our method (RCCycleGAN) is compared with CycleGAN and DerainCycleGAN in terms of image generation effects, and the performance of each model is quantitatively evaluated using the evaluation metrics in the previous section. CycleGAN and DerainCycleGAN are trained in the same experimental environment with the recommended parameter settings in the original paper.

Figure 8 displays the rain image generation effects of the above three models. In comparison with CycleGAN, RCCycleGAN performs better in terms of retaining image structural information, reducing image distortion, and having a lower level of image noise. In comparison with DerainCycleGAN, RCCycleGAN retains more color information, preserves more local details (shadows, reflections, etc.) in the image, and is closer to the real-world scenario.

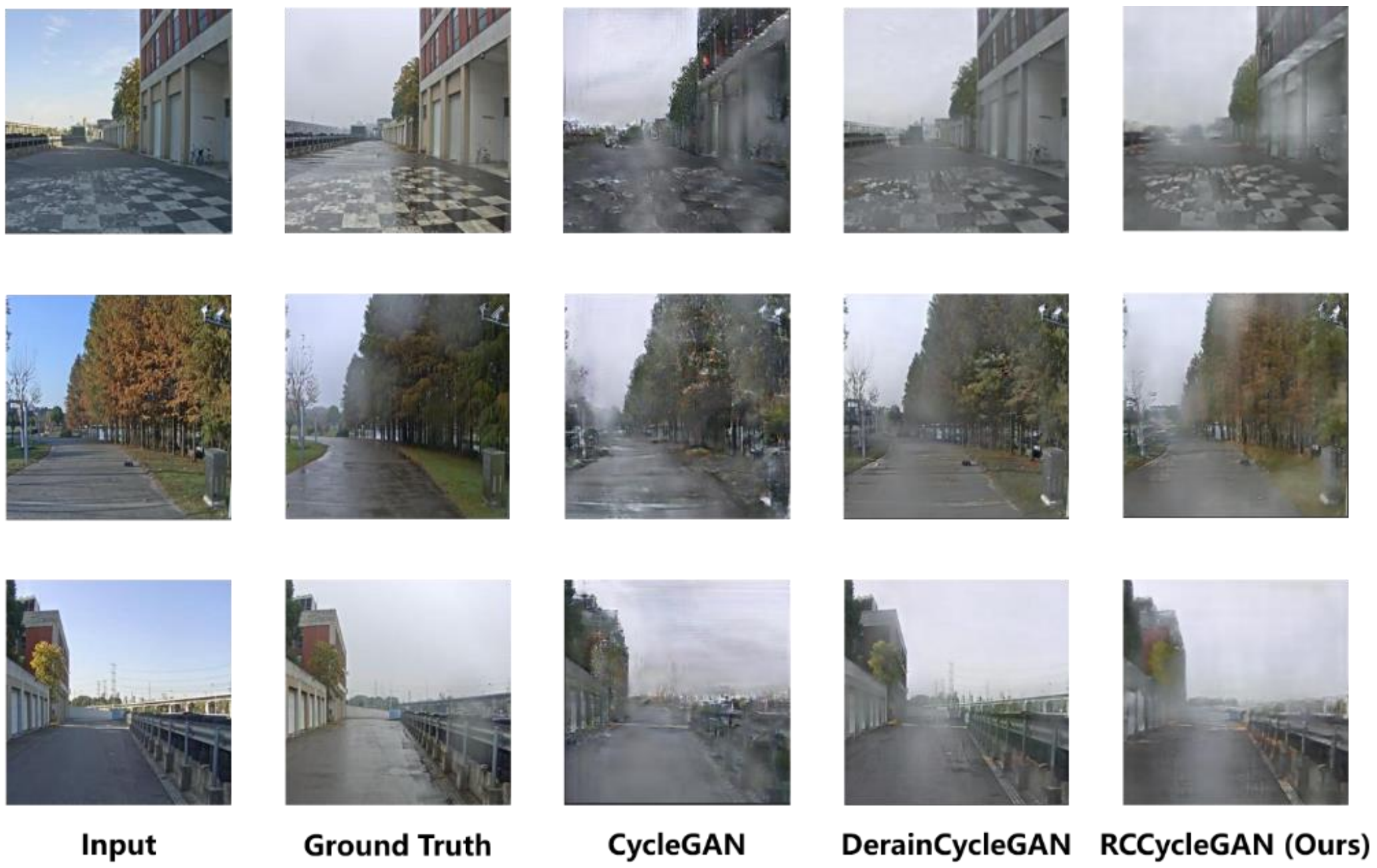

Figure 8 Model comparison

PSNR and SSIM values are calculated with the natural light rain image as the reference, and the results are shown in Figure 9. The PSNR and SSIM values of the rain images generated by RCCycleGAN are larger and the generated rain images are closer to the natural rain images consistent with the results of the subjective visual evaluation. Compared with CycleGAN, DerainCycleGAN has increased PSNR by 1.84dB and SSIM by 10%, while RCCycleGAN has increased PSNR by 0.74dB and SSIM by 8% compared with DerainCycleGAN.

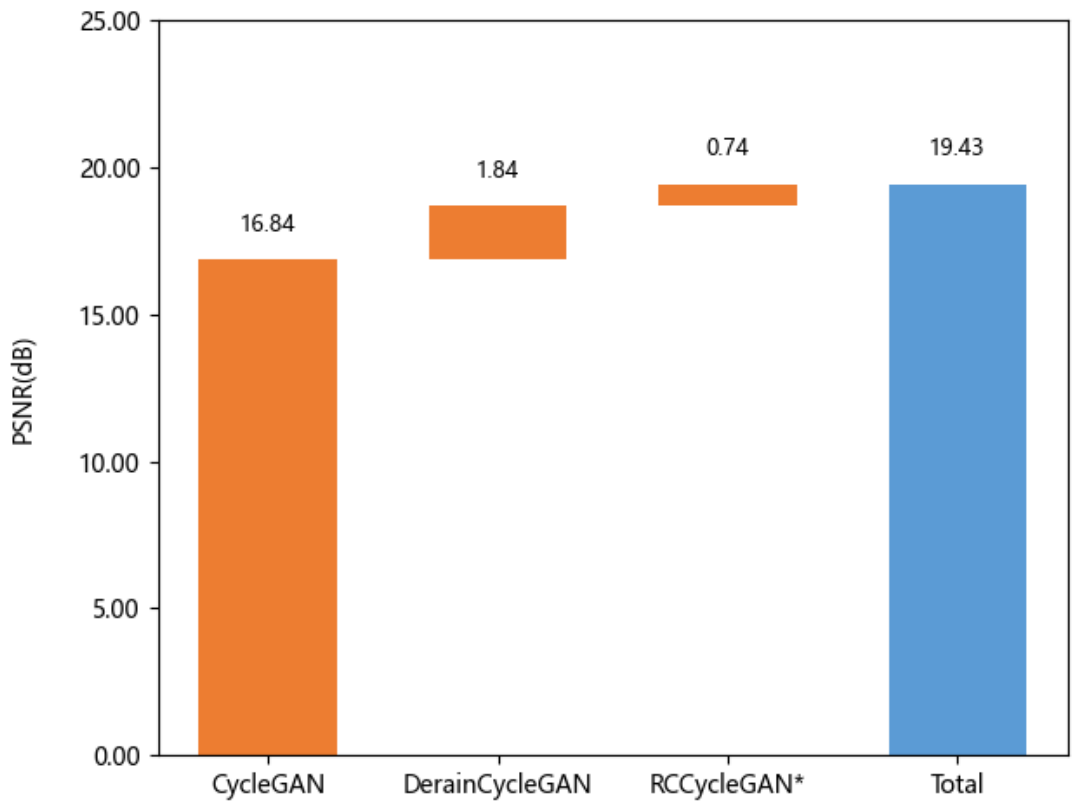

a) PSNR

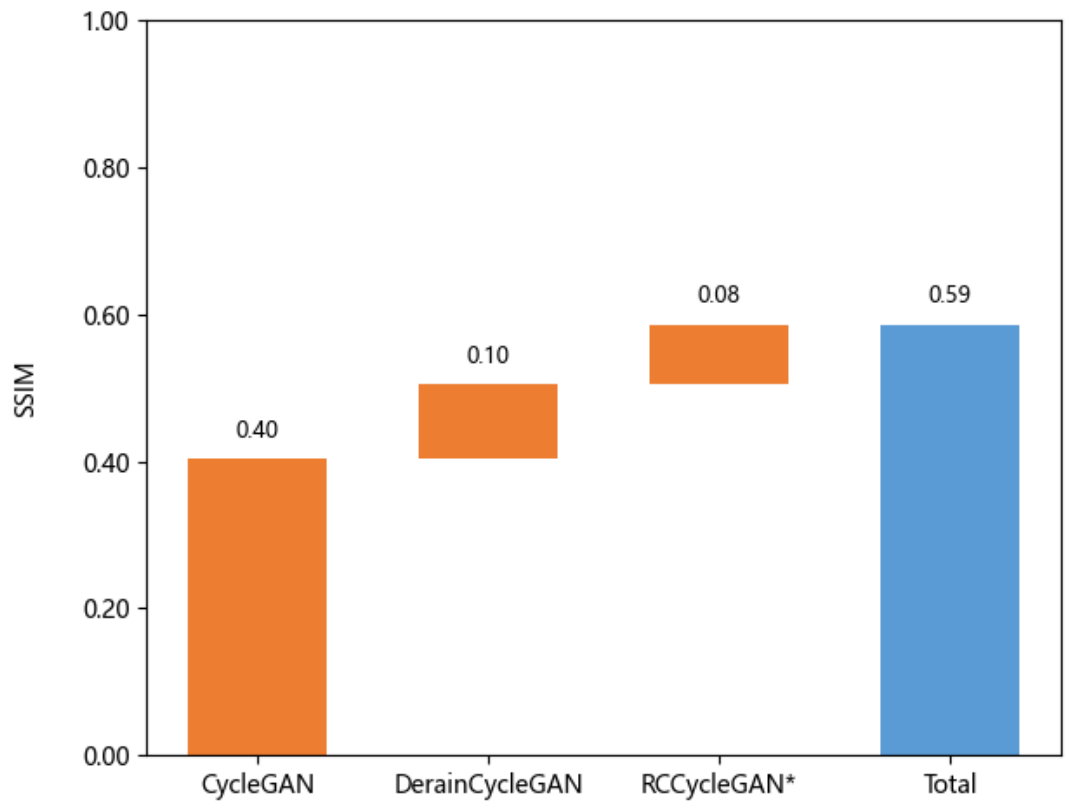

b) SSIM

Figure 9 PSNR and SSIM results

### 4.3 Ablation experiment

DerainCycleGAN is the basis of our model improvement. In the ideal case, the optimal performance is achieved by the adversarial training of the generator network and the discriminator network. However, because of the smaller optimization space of the discriminator network compared with the generator network, it converges quickly and then cannot guide the generator network to achieve diverse outputs, leading to mode collapse[33]. To deal with this problem, the original model is optimized in three aspects: 1) The labels of different rain intensities are introduced as the condition to achieve hierarchical-level rain image generation; 2) The discriminator network adopts the sigmoid activation function instead of the original LeakyReLU activation function to weaken its performance; 3) The suppression strategy is introduced to balance the dynamic game between the discriminator and generator networks. These enhancements are sequentially added to the baseline model, DerainCycleGAN, forming the final RCCycleGAN. This section conducts ablation experiments for each improvement and the results are shown in Figure 10.

The performance of the generator network significantly decreases after the introduction of rain intensity labels, as shown in Figure 10. PSNR decreases by 1.25dB and SSIM decreases by 10%. Subsequently, the performance of the overall model has gradually improved by weakening the performance of the discriminator network and adjusting the training strategy of the overall network. This suggests that for the optimal overall performance of the GAN, the discriminator network and the generator network need to be at a balanced adversarial level and converge together. A single network with superior performance cannot improve the overall performance of the network, and may even reduce it.

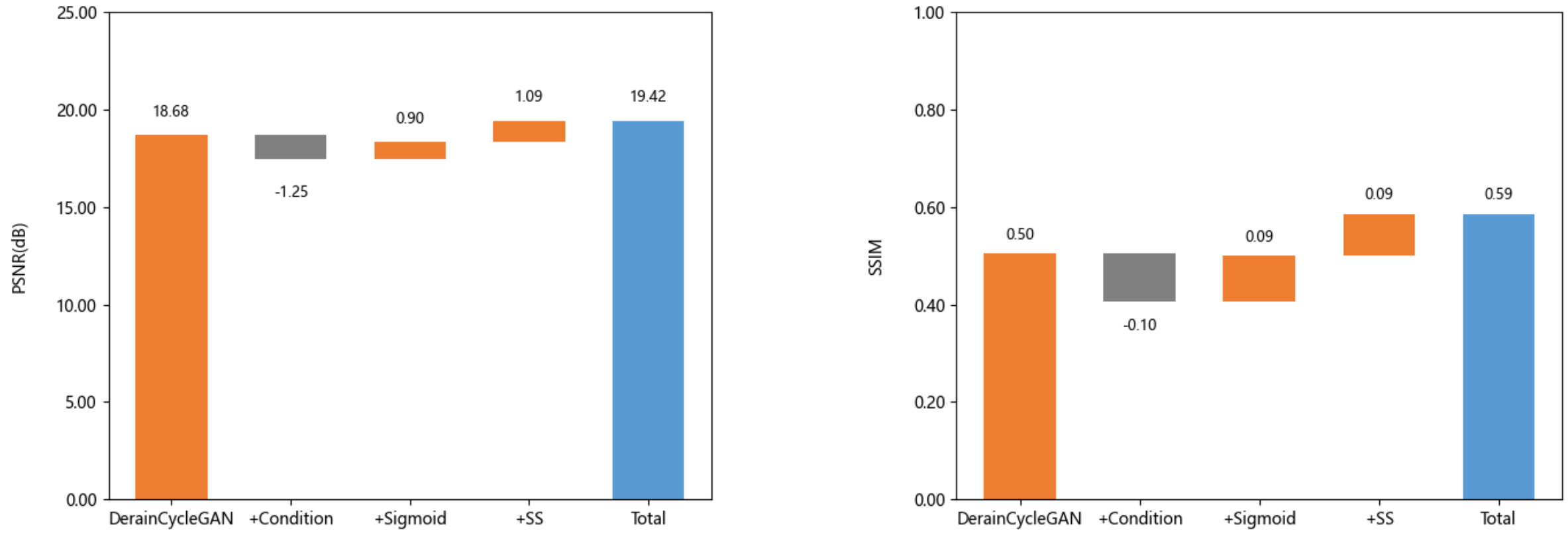

a) PSNR b) SSIM

Figure 10 The results of ablation experiments

## 5 Conclusion and outlook

In this paper, we construct a hierarchical-level rain image generative model to generate controllable rain intensity image. First, we investigate the existing image generation methods, among which GAN is widely used in the field of image generation as it can achieve optimal network performance through adversarial training without designing special loss functions. Then, we introduce the hierarchical-level rain image generative model (RCCycleGAN). The model and training strategy are improved to address the issue of excessive performance of the discriminator network. Next, we collect and produce a natural rain dataset, and also present the details of the training and the training effects. Finally, we verify the effectiveness of the model through the comparison with the baseline models. The results show that compared with CycleGAN and DerainCycleGAN, the PSNR of RCCycleGAN is improved by 2.58 dB and 0.74 dB, and the SSIM is improved by 18% and 8%, respectively. In addition, the results of ablation experiments verify the effectiveness of the model tuning.

Future research will focus on the following aspects:1) Evaluation of the model: In this paper, the performance of the model is assessed by image quality. In future research, object detection results will be added to verify the performance of the model. 2) Generalization of the model: The RCCycleGAN will be trained on existing public datasets to verify generalization.

## Reference

[1] WANG B, HAN Y, WANG S, et al. A Review of Intelligent Connected Vehicle Cooperative Driving Development[J]. Mathematics, 2022, 10(19): 3635.

[2] International Organization for Standardization. Road Vehicles-Functional Safety, ISO 26262 [S]. Geneva, Switzerland, 2011.

[3] International Organization for Standardization. Road Vehicles-Cybersecurity Engineering, Ground Vehicle Standard ISO/SAE 21434[S]. 2020.

[4] International Organization for Standardization. Road Vehicles-Safety of the Intended Functionality: ISO/DIS 21448-2021 [S]. Geneva, Switzerland: ISO, 2021: 1.

[5] Huang A, Xing X, Zhou T, et al. A Safety Analysis and Verification Framework for Autonomous Vehicles Based on the Identification of Triggering Events[R]. SAE Technical Paper, 2021.

[6] Jiang Wei, Xing Xingyu, Huang An, Chen Junyi. Research on Performance Limitations of Visual-based Perception System for Autonomous Vehicle under Severe Weather Conditions[C]//2022 IEEE Intelligent Vehicles Symposium (IV). IEEE, 2022: 586-593.

[7] Jiang Yuande, Sun Pengpeng, et al. Influence of Rain and Snow on Autonomous Vehicles' Camera[J]. China Journal of Highway and Transport,2022,35(03):307-316.

[8] Marathe A, Jain P, Walambe R, et al. RestoreX-AI: A Contrastive Approach towards Guiding Image Restoration via Explainable AI Systems[C]//Proceedings of the IEEE/CVF Conference on Computer Vision and Pattern Recognition. 2022: 3030-3039.

[9] Peng L, Wang H, Li J. Uncertainty evaluation of object detection algorithms for autonomous vehicles[J]. Automotive Innovation, 2021, 4(3): 241-252.

[10] Garg K, Nayar S K. Vision and rain[J]. International Journal of Computer Vision, 2007, 75(1): 3-27.

[11] Garg K, Nayar S K. Photorealistic rendering of rain streaks[J]. ACM Transactions on Graphics (TOG), 2006, 25(3): 996-1002.

[12] Barnum P C, Narasimhan S, Kanade T. Analysis of rain and snow in frequency space[J]. International journal of computer vision, 2010, 86(2): 256-274.

[13] Tremblay M, Halder S S, de Charette R, et al. Rain rendering for evaluating and improving robustness to bad weather[J]. International Journal of Computer Vision, 2021, 129(2): 341-360.

[14] Liu X, Suganuma M, Sun Z, et al. Dual residual networks leveraging the potential of paired operations for image restoration[C]//Proceedings of the IEEE/CVF Conference on Computer Vision and Pattern Recognition. 2019: 7007-7016.

[15] Zhu J Y, Park T, Isola P, et al. Unpaired image-to-image translation using cycle-consistent adversarial networks[C]//Proceedings of the IEEE international conference on computer vision. 2017: 2223-2232.

[16] Zhu H, Peng X, Zhou J T, et al. Singe image rain removal with unpaired information: A differentiable programming perspective[C]//Proceedings of the AAAI Conference on Artificial Intelligence. 2019, 33(01): 9332-9339.

[17] Goodfellow I, Pouget-Abadie J, Mirza M, et al. Generative adversarial networks[J]. Communications of the ACM, 2020, 63(11): 139-144.

[18] Mirza M, Osindero S. Conditional generative adversarial nets[J]. arXiv preprint arXiv:1411.1784, 2014.

[19] Radford A, Metz L, Chintala S. Unsupervised representation learning with deep convolutional generative adversarial networks[J]. arXiv preprint arXiv:1511.06434, 2015.

[20] Zhu J Y, Park T, Isola P, et al. Unpaired image-to-image translation using cycle-consistent adversarial networks[C]//Proceedings of the IEEE international conference on computer vision. 2017: 2223-2232.

[21] Wei Y, Zhang Z, Wang Y, et al. Deraincyclegan: Rain attentive cyclegan for single image deraining and rainmaking[J]. IEEE Transactions on Image Processing, 2021, 30: 4788-4801.

[22] Shi X, Chen Z, Wang H, et al. Convolutional LSTM network: A machine learning approach for precipitation nowcasting[J]. Advances in neural information processing systems, 2015, 28.

[23] Geiger A, Lenz P, Stiller C, et al. Vision meets robotics: The kitti dataset[J]. The International Journal of Robotics Research, 2013, 32(11): 1231-1237.

[24] Caesar H, Bankiti V, Lang A H, et al. nuscenes: A multimodal dataset for autonomous driving[C]//Proceedings of the IEEE/CVF conference on computer vision and pattern recognition. 2020: 11621-11631.

[25] Yu F, Chen H, Wang X, et al. Bdd100k: A diverse driving dataset for heterogeneous multitask learning[C]//Proceedings of the IEEE/CVF conference on computer vision and pattern recognition. 2020: 2636-2645.

[26] Cordts M, Omran M, Ramos S, et al. The cityscapes dataset for semantic urban scene understanding[C]//Proceedings of the IEEE conference on computer vision and pattern recognition. 2016: 3213-3223.

[27] Pitropov M, Garcia D E, Rebello J, et al. Canadian adverse driving conditions dataset[J]. The International Journal of Robotics Research, 2021, 40(4-5): 681-690.

[28] Burnett K, Yoon D J, Wu Y, et al. Boreas: A multi-season autonomous driving dataset[J]. The International Journal of Robotics Research, 2023, 42(1-2): 33-42.

[29] Wang T, Yang X, Xu K, et al. Spatial attentive single-image deraining with a high quality real rain dataset[C]//Proceedings of the IEEE/CVF Conference on Computer Vision and Pattern Recognition. 2019: 12270-12279.

[30] Wei W, Meng D, Zhao Q, et al. Semi-supervised transfer learning for image rain removal[C]//Proceedings of the IEEE/CVF conference on computer vision and pattern recognition. 2019: 3877-3886.

[31] Zhang H, Sindagi V, Patel V M. Image de-raining using a conditional generative adversarial network[J]. IEEE transactions on circuits and systems for video technology, 2019, 30(11): 3943-3956.

[32] Fu X, Huang J, Zeng D, et al. Removing rain from single images via a deep detail network[C]//Proceedings of the IEEE conference on computer vision and pattern recognition. 2017: 3855-3863.

[33] Bau D, Zhu J Y, Wulff J, et al. Seeing what a gan cannot generate[C]. Proceedings of the IEEE/CVF International Conference on Computer Vision. 2019: 4502-4511.